\def\year{2020}\relax
\documentclass[letterpaper]{article} % DO NOT CHANGE THIS
\usepackage{aaai20}  % DO NOT CHANGE THIS
\usepackage{times}  % DO NOT CHANGE THIS
\usepackage{helvet} % DO NOT CHANGE THIS
\usepackage{courier}  % DO NOT CHANGE THIS
\usepackage[hyphens]{url}  % DO NOT CHANGE THIS
\usepackage{graphicx} % DO NOT CHANGE THIS
\usepackage{graphicx}  % DO NOT CHANGE THIS
\usepackage{array}
\usepackage{amsmath}
\usepackage{amssymb}
\usepackage{booktabs}
\usepackage{bm}

\title{Neural Cognitive Diagnosis for Intelligent Education Systems}
\author{
	Fei Wang\textsuperscript{\rm 1}, Qi Liu\textsuperscript{\rm 1*}, Enhong Chen\textsuperscript{\rm 1}\thanks{Corresponding Authors.}, Zhenya Huang\textsuperscript{\rm 1} \\ \Large \textbf{Yuying Chen\textsuperscript{\rm 1}, Yu Yin\textsuperscript{\rm 1}, Zai Huang\textsuperscript{\rm 1}, Shijin Wang\textsuperscript{\rm 2}}\\ \textsuperscript{\rm 1}Anhui Province Key Lab. of Big Data Analysis and Application, \\School of Computer Science and Technology, University of Science and Technology of China\\ \textsuperscript{\rm 2}iFLYTEK Research\\ wf314159@mail.ustc.edu.cn, \{qiliuql, cheneh\}@ustc.edu.cn,\\ \{huangzhy, cyy33222, yxonic, huangzai\}@mail.ustc.edu.cn, sjwang3@iflytek.com
}

\begin{document}

\maketitle

\begin{abstract}
	Cognitive diagnosis is a fundamental issue in intelligent education, which aims to discover the proficiency level of students on specific knowledge concepts. Existing approaches usually mine linear interactions of student exercising process by manual-designed function (e.g., logistic function), which is not sufficient for capturing complex relations between students and exercises. In this paper, we propose a general Neural Cognitive Diagnosis (NeuralCD) framework, which incorporates neural networks to learn the complex exercising interactions, for getting both accurate and interpretable diagnosis results. Specifically, we project students and exercises to factor vectors and leverage multi neural layers for modeling their interactions, where the monotonicity assumption is applied to ensure the interpretability of both factors. Furthermore, we propose two implementations of NeuralCD by specializing the required concepts of each exercise, i.e., the NeuralCDM with traditional Q-matrix and the improved NeuralCDM+ exploring the rich text content. Extensive experimental results on real-world datasets show the effectiveness of NeuralCD framework with both accuracy and interpretability.
	%Cognitive diagnosis in intelligent education aims to discover the proficiency level of students on specific knowledge concepts. Existing approaches usually mine linear interactions of student exercising process by manually designed function, which is not sufficient for capturing complex relations between students and exercises. In this paper, we propose a general Neural Cognitive Diagnosis (NeuralCD) framework, which incorporates neural networks to learn the complex interactions between student's and exercise's factor vectors. To ensure the interpretability of factor vectors, which is significant for diagnosis, we apply the monotonicity assumption borrowed from educational psychology. A specific model NeuralCDM is provided as an implementation example of the framework. We further improve NeuralCDM by exploring the text content to show the extendability of NeuralCD, and prove how NeuralCD covers some traditional diagnostic models to demonstrate its generality. Extensive experimental results on real-world datasets show the effectiveness of NeuralCD framework with both accuracy and interpretability.	
\end{abstract}

\section{Introduction}
\begin{figure*}[h]
	\centering
	\includegraphics[scale=0.63]{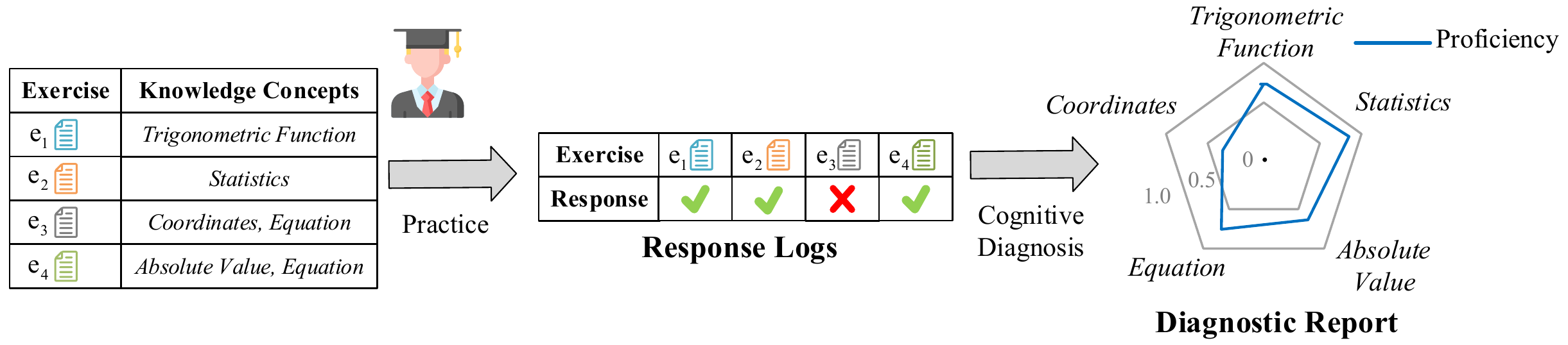}
	\caption{A toy example of cognitive diagnosis. The student choose some exercises for practice and leave the response logs. With cognitive diagnosis methods, we get the diagnostic report containing the student's proficiency on each knowledge concept.}
	\label{fig:diagnose_example}
\end{figure*}

Cognitive diagnosis is a necessary and fundamental task in many real-world scenarios such as games~\cite{chen2016predicting}, medical diagnosis~\cite{guo2017modeling}, and education. Specifically, in intelligent education systems~\cite{anderson2014engaging,burns2014intelligent}, cognitive diagnosis aims to discover the states of students in the learning process, such as their proficiencies on specific knowledge concepts~\cite{liu2018fuzzy}. Figure~\ref{fig:diagnose_example} shows a toy example of cognitive diagnosis. Generally, students usually first choose to practice a set of exercises (e.g., $e_1, \cdots, e_4$) and leave their responses (e.g., right or wrong). Then, our goal is to infer their actual knowledge states on the corresponding concepts (e.g., {\it Trigonometric Function}). In practice, these diagnostic reports are necessary as they are the basis of further services, such as exercise recommendation and targeted training~\cite{kuh2011piecing}.

In the literature, massive efforts have been devoted for cognitive diagnosis, such as Deterministic Inputs, Noisy-And gate model (DINA)~\cite{de2009dina}, Item Response Theory (IRT)~\cite{embretson2013item}, Multidimensional IRT (MIRT)~\cite{reckase2009multidimensional} and Matrix Factorization (MF)~\cite{koren2009matrix}. Despite achieving some effectiveness, these works rely on handcrafted interaction functions that just combine the multiplication of student's and exercise's trait features linearly, such as logistic function~\cite{embretson2013item} or inner product~\cite{koren2009matrix}, which may not be sufficient for capturing the complex relationship between students and exercises~\cite{dibello200631a}. Besides, the design of specific interaction functions is also labor-intensive since it usually requires professional expertise. Therefore, it is urgent to find an automatic way to learn the complex interactions for cognitive diagnosis instead of manually designing them.

In this paper, we address this issue in a principled way of proposing a Neural Cognitive Diagnosis (NeuralCD) framework by incorporating neural networks to model complex non-linear interactions. Although the capability of neural networks to approximate continuous functions has been proved in many domains, such as natural language processing~\cite{zhang2018navigating} and recommender systems~\cite{song2019hierarchical}, it is still highly nontrivial to adapt to cognitive diagnosis due to the following domain challenges. First, the black-box nature of neural networks makes them difficult to get explainable diagnosis results. That is to say, it is difficult to explicitly realize how much a student has mastered a certain knowledge concept (e.g., {\it Equation}). Second, traditional models are designed manually with non-neural functions, which makes it hard for them to leverage exercise text content. However, with neural network, it is worthy of finding ways to explore the rich information contained in exercise text content for cognitive diagnosis.

To address these challenges, we propose a NeuralCD framework to approximate interactions between students and exercises, yet preserving the explainability. We first project students and exercises to factor vectors and leverage multi-layers for modeling the complex interactions of student answering exercises. To ensure the interpretability of both factors, we apply the monotonicity assumption taking from educational property~\cite{reckase2009multidimensional} on the multi-layers. Then, we propose two implementations on the basis of the general framework, i.e., NeuralCDM and NeuralCDM+. In NeuralCDM, we simply extract exercise factor vectors from traditional Q-matrix (an example is shown in Figure~\ref{fig:case_study}) and achieve the monotonicity property with positive full connection layers, which shows feasibility of the framework. While in NeuralCDM+, we demonstrate how information from exercise text can be explored with neural network to extend the framework. Particularly, our NeuralCD is a general framework since it can cover many traditional models such as MF, IRT and MIRT. Finally, we conduct extensive experiments on real-world datasets, and the results show the effectiveness of NeuralCD framework with both accuracy and interpretability guarantee.

Our code of NeuralCDM is available at https://github.com/bigdata-ustc/NeuralCD.

\section{Related Work}
In this section, we briefly review the related works from the following three aspects.
%   提工作时要用过去式

\noindent\textbf{Cognitive Diagnosis.} Existing works about student cognitive diagnosis mainly came from educational psychology area. DINA~\cite{de2009dina,von2014dina} and IRT~\cite{embretson2013item} were two of the most typical works, which model the result of a student answering an exercise as the interaction between the trait features of the student ($\bm \theta$) and the exercise ($\bm \beta$). Specifically, in DINA, $\bm \theta$ and $\bm \beta$ were binary, where $\bm \beta$ came directly from Q-matrix (a human labeled exercise-knowledge correlation matrix, an example is showed in Figure~\ref{fig:case_study}). Another two exercise factors, i.e. guessing and slipping (parameterized as $g$ and $s$) are also taken into consideration. The probability of student $i$ correctly answering exercise $j$ was modeled as $P(r_{ij}=1|{\bm \theta_i})=g_j^{1-\eta_{ij}}(1-s_j)^{\eta_{ij}}$, where $\eta_{ij}=\prod_{k}\theta_{ik}^{\beta_{jk}}$. On the other hand, in IRT, $\bm \theta$ and $\bm \beta$ were unidimensional and continuous latent traits, indicating student ability and exercise difficulty respectively. The interaction between the trait features was modeled in a logistic way, e.g., a simple version is $sigmoid(a(\bm \theta - \bm \beta))$, where $a$ is the exercise discrimination parameter. Although extra parameters were added in IRT~\cite{fischer1995derivations,lord2012applications} and latent trait was extended to multidimensional(MIRT)~\cite{adams1997multidimensional,reckase2009multidimensional}, most of their item response functions were still logistic-like. These traditional models depended on manually designed functions, which was labor-intensive and restricted their scope of applications. 

\noindent\textbf{Matrix Factorization.} Recently, some researches from data mining perspective have demonstrated the feasibility of MF for cognitive diagnosis. Student and exercise correspond to user and item in matrix factorization (MF). For instance, Toscher et al.~\shortcite{toscher2010collaborative} improved SVD (Singular Value Decomposition) methods to factor the score matrix and get students and exercises' latent trait vectors. Thai-Nghe et al.~\shortcite{thai2010recommender} applied some recommender system techniques including matrix factorization in the educational context, and compared it with traditional regression methods. Besides, Thai-Nghe et al.~\shortcite{thai2015multi} proposed a multi-relational factorization approach for student modeling in the intelligent tutoring systems. Despite their effectiveness in predicting students' scores on exercises, the latent trait vectors in MF is not interpretable for cognitive diagnosis, i.e. there is no clear correspondence between elements in trait vectors and specific knowledge concepts.

\noindent\textbf{Artificial Neural Network.} Techniques using artificial neural network have reached state-of-the-art in many areas, e.g., speech recognition~\cite{chan2016listen}, text classification~\cite{zhang2015character} and image captioning~\cite{wang2019hierarchical}. There are also some educational applications such as question difficulty prediction~\cite{huang2017question}, code education~\cite{wu2019zero}, formula image transcribing~\cite{yin2018transcribing} and student performance prediction~\cite{huang2019ekt}. However, using neural network for cognitive diagnosis is nontrivial as it performs poorly in parameter interpretation due to its inherent traits. To the best of our knowledge, deep knowledge tracing (DKT)~\cite{piech2015deep} was the first attempt to model student learning process using recurrent neural network. However, DKT aims to predict students' scores, and does not make a distinction between an exercise and the knowledge concepts it contains, thus it's unsuitable for cognitive diagnosis. Few works with neural network have high interpretability for student cognitive diagnosis. Towards this end, in this paper we propose a neural cognitive diagnosis (NeuralCD) framework which borrows concepts from educational psychology and combine them with interaction functions learned from data. NeuralCD could achieve both high accuracy and interpretation with neural network. Besides, the framework is general that can cover many tradition models, and at the same time easy for extension.

\section{Neural Cognitive Diagnosis}
We first formally introduce cognitive diagnosis task. Then we describe the details of NeuralCD framework. After that, we design a specific diagnostic network NeuralCDM with traditional Q-matrix to show the feasibility of the framework, and an improved NeuralCDM+ by incorporating exercise text content for better performance. Finally, we demonstrate the generality of NeuralCD framework by showing its close relationship with some traditional models.

\subsection{Task Overview}
Suppose there are $N$ Students, $M$ Exercises and $K$ Knowledge concepts at a learning system, which can be represented as $S=\{s_1, s_2, \dots, s_N\}, E=\{e_1, e_2, \dots, e_M\}$ and $K_n=\{k_1, k_2,\dots, k_K\}$ respectively. Each student will choose some exercises for practice, and the response logs $R$ are denoted as set of triplet $(s, e, r)$ where $s \in S, e \in E$ and $r$ is the score (transferred to percentage) that student $s$ got on exercise $e$. In addition, we have Q-matrix (usually labeled by experts) $\mathbf{Q}=\{Q_{ij}\}_{M \times K}$, where $Q_{ij}=1$ if exercise $e_i$ relates to knowledge concept $k_j$ and $Q_{ij}=0$ otherwise.

\textbf{Problem Definition} {\it Given students' response logs $R$ and the Q-matrix $\mathbf Q$, the goal of our cognitive diagnosis task is to mine students' proficiency on knowledge concepts through the student performance prediction process.}

\subsection{Neural Cognitive Diagnosis Framework}
\begin{figure}
	\centering
	\includegraphics[scale=0.55]{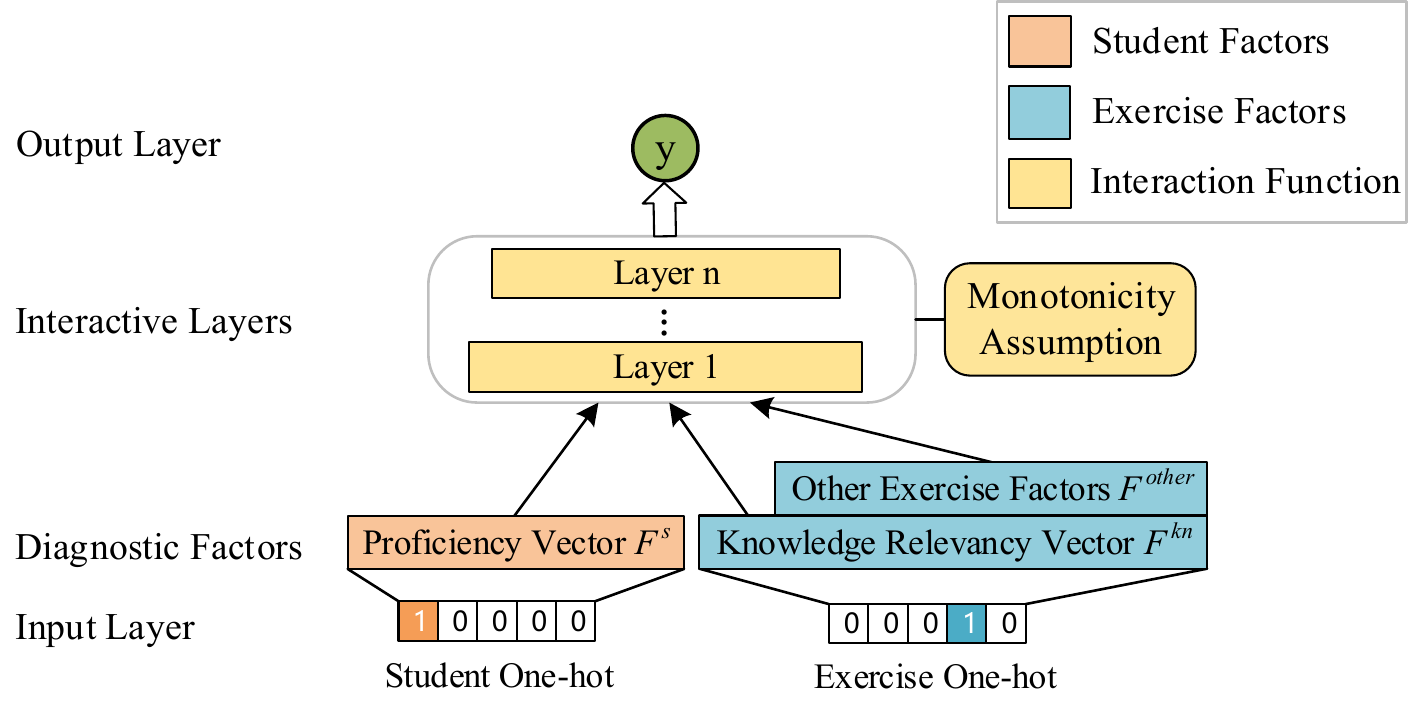}
	\caption{Structure of NeuralCD framework.}
	\label{fig:framework}
\end{figure}

Generally, for a cognitive diagnostic system, there are three elements need to be considered: student factors, exercise factors and the interaction function among them~\cite{dibello200631a}. In this paper, we propose a general NeuralCD framework to address them by using multi-layer neural network modeling, which is shown in Figure~\ref{fig:framework}. Specifically, for each response log, we use one-hot vectors of the corresponding student and exercise as input and obtain the diagnostic factors of the student and exercise. Then the interactive layers learn the interaction function among the factors and output the probability of correctly answering the exercise. After training, we get students' proficiency vectors as diagnostic results. Details are introduced as bellow.

\paragraph{Student Factors.} Student factors characterize the traits of students, which would affect the students' response to exercises. As our goal is to mine students' proficiency on knowledge concepts, we do not use the latent trait vectors as in IRT and MIRT, which is not explainable enough to guide students' self-assessment. Instead, we design the student factors as explainable vectors similar to DINA, but has a major difference that they are continuous. Specifically, We use a vector $F^s$ to characterize a student, namely {\it proficiency vector}. Each entry of $F^s$ is continuous ([0,1]), which indicates the student's proficiency on a knowledge concept. For example, $F^s=[0.9,0.2]$ indicates a high mastery on the first knowledge concept but low mastery on the second. $F^s$ is got through the parameter estimation process.

\paragraph{Exercise Factors.} Exercise factors denote the factors that characterize the traits of exercises. We divide exercise factors into two categories. The first indicates the relationship between exercises and knowledge concepts, which is fundamental as we need it to make each entry of $F^s$ correspond to a specific knowledge concept for our diagnosis goal. We call it {\it knowledge relevancy vector} and denote it as $F^{kn}$. $F^{kn}$ has the same dimension as $F^s$, with the $i$th entry indicating the relevancy between the exercise and the knowledge concept $k_i$. Each entry of $F^{kn}$ is non-negative. $F^{kn}$ is previously given (e.g., obtained from Q-matrix). Other factors are of the second type and are optional. Factors from IRT and DINA such as knowledge difficulty, exercise difficulty and discrimination can be incorporated if reasonable.

\paragraph{Interaction Function.} We use artificial neural network to obtain the interaction function for the following reasons. First, the neural network has been proven to be capable of approximating any continuous function~\cite{hornik1989multilayer}. The strong fitting ability of neural network makes it competent for capturing relationships among student and exercise factors. Second, with neural network, the interaction function can be learned from data with few assumptions (that behind traditional models). This makes NeuralCD more general and can be applied in broad areas. Third, the framework can be highly extendable with neural network. For instance, extra information such as exercise texts can be integrated in with neural network (We will discuss its extendability in the following subsections.). Mathematically, we formulate the output of NeuralCD framework as:
\begin{equation}
y=\varphi_n(\dots\varphi_1(F^s, F^{kn}, F^{other}, \theta_f)),
\end{equation}
where $\varphi_i$ denotes the mapping function of the $i$th MLP layer; $F^{other}$ denotes factors other than $F^s$ and $F^{kn}$ (e.g., difficulty); and $\theta_f$ denotes model parameters of all the interactive layers.

However, due to some intrinsic characteristics, neural networks usually have poor performance on interpretation~\cite{samek2016evaluating}. Fortunately, we find that the monotonicity assumption, which is used in some IRT and MIRT models~\cite{reckase2009multidimensional}, can be utilized to ensure the interpretation of student and exercise factors. Monotonicity assumption is general and reasonable in almost all circumstance, thus it has little influence on the generality of NeuralCD framework. The assumption is defined as follows:
\begin{figure*}[ht]
	\begin{minipage}[t]{0.47\textwidth}
		\centering
		\includegraphics[scale=0.6]{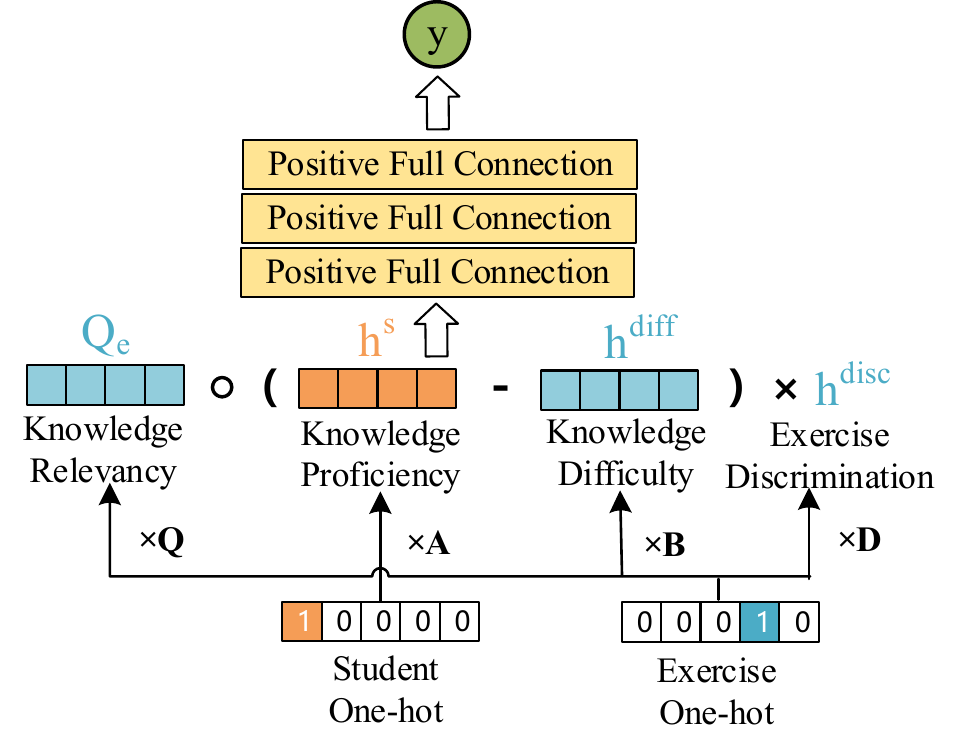}
		\caption{Neural cognitive diagnosis model. The color of orange, blue and yellow indicate student factors, exercise factors and interacion function respectively.}
		\label{fig:NeuralCDM}
	\end{minipage}\hspace{9mm}
	\begin{minipage}[t]{0.47\textwidth}
		\centering
		\includegraphics[scale=0.51]{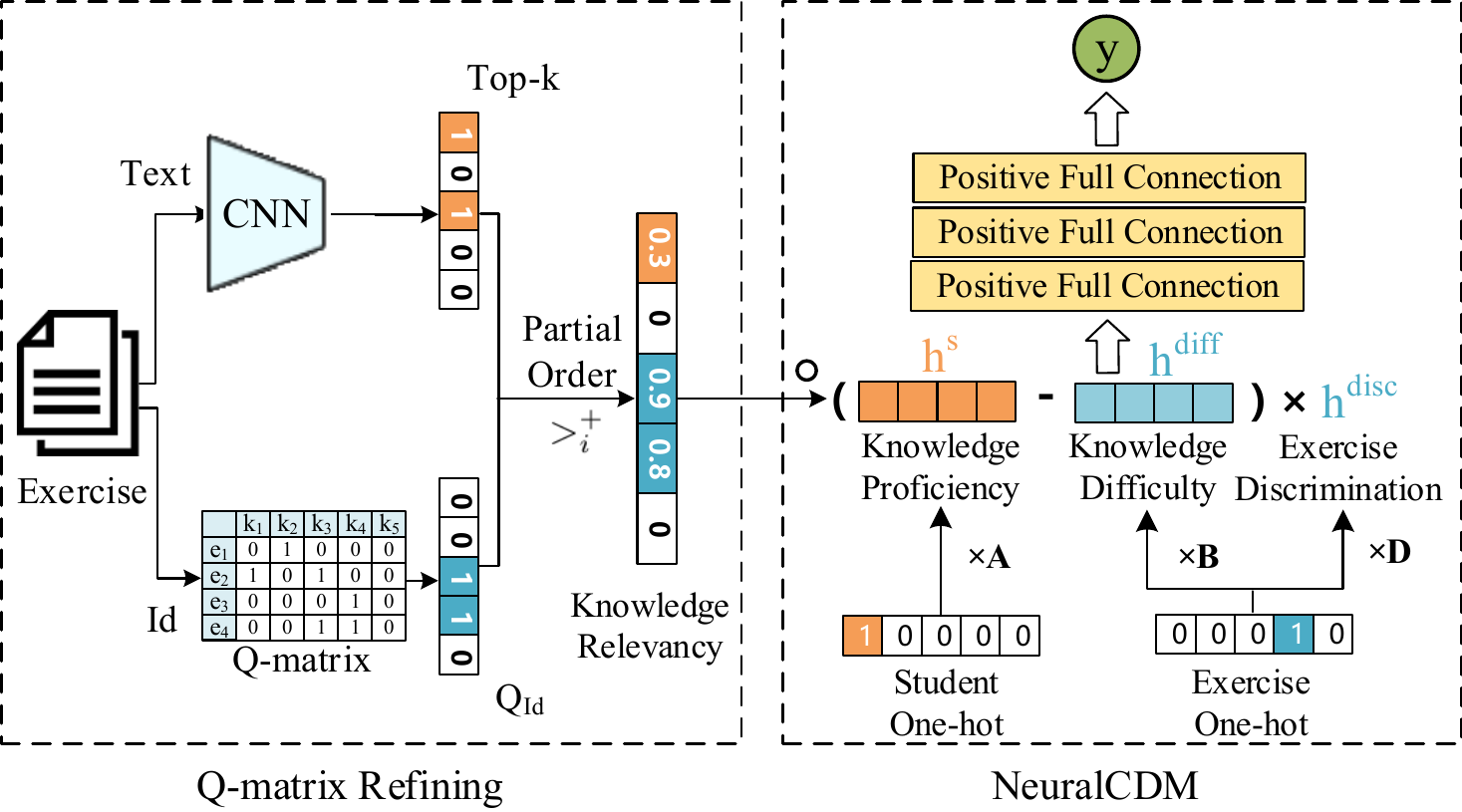}
		\caption{Extended neural cognitive diagnosis model. The knowledge relevancy vectors are replaced with vectors in Q-matrix that is refined by leveraging exercise texts.}
		\label{fig:NeuralCDM+}
	\end{minipage}
\end{figure*}

\textbf{Monotonicity Assumption} {\it The probability of correct response to the exercise is monotonically increasing at any dimension of the student's knowledge proficiency.}

This assumption should be converted as a property of the interaction function. Intuitively, we assume student $s$ to answer exercise $e$ correctly. During training, the optimization algorithm should increase the student's proficiency if the model output a wrong prediction (i.e., a value below 0.5). The increment of each knowledge proficiency is otherwise controlled by $F^{kn}$.  

After introducing the structure of NeuralCD framework, we will next show some specific implementations. We first design a diagnostic model based on NeuralCD with extra exercise factors (i.e., knowledge difficulty and exercise discrimination)($\S$\ref{NeuralCDM}), and further show its extendability by incorporating text information ($\S$\ref{NeuralCDM+}) and generality by demonstrating how it covers traditional models ($\S$\ref{Generality}).

\subsection{Neural Cognitive Diagnosis Model} \label{NeuralCDM}
Here we introduce a specific neural cognitive diagnosis model (NeuralCDM) under NeuralCD framework. Figure~\ref{fig:NeuralCDM} illustrates the structure of NeuralCDM.

\noindent\textbf{Student Factors.} In NeuralCDM, each student is represented with a knowledge proficiency vector. The student factor $F^s$ aforementioned is $\bm{h}^s$ here, and $\bm{h}^s$ is obtained by multiplying the student's one-hot representation vector $\bm{x}^s$ with a trainable matrix $\mathbf{A}$. That is,
\begin{equation}
\bm{h}^s={\rm sigmoid}(\bm{x}^s\times \mathbf{A} ), 
\end{equation}
in which $\bm{h}^s \in (0,1)^{1\times K}, \bm{x}^s\in \{0,1\}^{1\times N}, \mathbf{A}\in \mathbb{R}^{N\times K}$.

\noindent\textbf{Exercise Factors.} As for each exercise, the aforementioned exercise factor $F^{kn}$ is $\bm{Q}_e$ here, which directly comes from the pre-given Q-matrix:
\begin{equation}
\bm{Q}_e=\bm{x}^e \times \mathbf{Q},
\end{equation}
where $\bm{Q}_e \in \{0, 1\}^{1 \times K}$, $\bm{x}^e \in \{0, 1\}^{1 \times M}$ is the one-hot representation of the exercise.
In order to make a more precise diagnosis, we adopt other two exercise factors: knowledge difficulty $\bm{h}^{diff}$ and exercise discrimination $h^{disc}$. $\bm{h}^{diff} \in (0, 1)^{1 \times K}$, indicates the difficulty of each knowledge concept examined by the exercise, which is extended from exercise difficulty used in IRT. $h^{disc} \in (0, 1)$, used in some IRT and MIRT models, indicates the capability of the exercise to differentiate between those students whose knowledge mastery is high from those with low knowledge mastery. They can be obtained by:
\begin{align}
\bm{h}^{diff}={\rm sigmoid}(\bm{x}^e \times \mathbf{B} ), \mathbf{B} \in \mathbb{R}^{M\times K} \\
h^{disc}={\rm sigmoid}(\bm{x}^e \times \mathbf{D}), \mathbf{D}\in \mathbb{R}^{M\times 1}
\end{align}
where $\mathbf{B}$ and $\mathbf{D}$ are trainable matrices.

\noindent\textbf{Interaction Function.} The first layer of the interaction layers is inspired by MIRT models. We formulate it as:
\begin{equation}
\bm{x} = \bm{Q}_e \circ (\bm{h}^s - \bm{h}^{diff}) \times h^{disc},
\label{eq:first_layer}
\end{equation}
where $\circ$ is element-wise product. Following are two full connection layers and an output layer:
\begin{align}
\bm{f}_1=\phi(\mathbf{W}_1 \times \bm{x}^T + \bm{b}_1),
\label{eq:full_layer1}\\
\bm{f}_2=\phi(\mathbf{W}_2 \times \bm{f}_1 + \bm{b}_2),
\label{eq:full_layer2}\\
y=\phi(\mathbf{W}_3 \times \bm{f}_2 + b_3),
\label{eq:output_layer}
\end{align}
where $\phi$ is the activation function. Here we use Sigmoid.

Different methods can be used to satisfy the monotonicity assumption. We adopt a simple strategy: restrict each element of $\mathbf{W}_1, \mathbf{W}_2, \mathbf{W}_3$ to be positive. It can be easily proved that $\frac{\partial y}{\partial h_i^s}$ is positive for each entry $h_i^s$ in $\bm{h}^s$. Thus monotonicity assumption is always satisfied during training.

The loss function of NeuralCDM is cross entropy between output $y$ and true label $r$:
\begin{equation}
loss_{CDM} =-\sum_{i}(r_i\log y_i + (1-r_i)\log (1-y_i)).
\end{equation}

After training, the value of $\bm{h}^s$ is what we get as diagnosis result, which denotes the student's knowledge proficiency. 

\subsection{NeuralCD Extension with Text Information}\label{NeuralCDM+}
In NeuralCDM and some traditional methods (e.g., DINA), Q-matrix is the source of information about the exercise knowledge concept. However, manually-labeled Q-matrix may be deficient because of inevitable errors and subjective bias~\cite{liu2012data,dibello200631a}. On the other hand, exercise texts have been proved to be highly related to some exercise features (e.g., difficulty, relevant knowledge concepts)~\cite{su2018exercise,huang2017question}, thus it can be leveraged to refine the Q-matrix. For example, in Q-matrix, maybe only '{\it Equation}' is labeled for an equation solving exercise. However, we may discover that '{\it Division}' is also required due to the existence of '$\div$' in the text. Traditional cognitive models didn't leverage text content due to the limitation of their handcraft non-neural interaction functions. However, with neural network, we are able to incorporate text information into our framework. We denote the extended model as NeuralCDM+, and present its structure in Figure~\ref{fig:NeuralCDM+}.
% We now show the extendability of NerualCD through the use of exercise texts. In traditional methods, exercise texts are not used for modeling. However, these texts contain important information about the exercises which can be useful for diagnosis, such as exercise difficulty and related knowledge concepts. Here we use exercise texts to find possible relevant knowledge concepts, and use them to refine manually-labeled Q-matrix, which is deficient because of inevitable errors and subjective bias~\cite{liu2012data,dibello200631a}. For example, in Q-matrix, maybe only '{\it Equation}' is labeled for an equation solving exercise. However, we may discover that '{\it Division}' is also required due to the existence of '$\div$' in the text. We denote the extended model as NeuralCDM+, and present its structure in Figure~\ref{fig:NeuralCDM+}.

Specifically, we first pre-train a CNN (convolutional neural network) to predict knowledge concepts related to the input exercise. CNN has advantage of extracting local information in text processing, thus it's able to capture important words from texts (e.g., words that are highly relative to certain knowledge concepts). The network takes concatenated word2vec embedding of words in texts as input, and output the relevancy of each predefined knowledge concept (that has occurred in data) to the exercise. Human-labeled Q-matrix is used as label for training. We define $V^k_i=\{V_{ij_1}, V_{ij_2}, \dots, V_{ij_k}\}$ as the set of top-k knowledge concepts of exercise $e_i$ outputted by the CNN. 

Then we combine $V^k_i$ with Q-matrix. Although there are defects in human-labeled Q-matrix, it still has high confidence. Thus we consider knowledge concepts labeled by Q-matrix are more relative than concepts in $\{k_j|k_j \in V^k_i \ and\ Q_{ij}=0 \}$. To achieve this, we adopt a pairwise Bayesian method as follows. For convenience, we define partial order $>_i^+$ as:
\begin{equation}
a >^+_i b,\ \text{if}\ Q_{ia}=1\ \text{and}\ Q_{ib}=0\ \text{and}\ b\in V^k_i,
\end{equation}
and define the partial order relationship set as $D_V=\{(i,a,b)|a >^+_i \ b, i=1, 2, \dots, M\}$.
Following traditional Bayesian treatment, we assume $\tilde{\mathbf{Q}}$ follows a zero mean Gaussian prior with standard deviation $\sigma$ of each dimension. To give Q-matrix labels higher confidence, we define $p(a >^+_i b|\tilde{\mathbf{Q}}_i)$ with a pairwise logistic-like function:
\begin{equation}
p(a >^+_i b|\tilde{\bm{Q}}_i)=\frac{1}{1+e^{-\lambda(\tilde{Q}_{ia}-\tilde{Q}_{ib})}}.
\label{eq:q_logistic}
\end{equation}
The parameter $\lambda$ controls the discrimination of relevance values between labeled and unlabeled knowledge concepts.
The log posterior distribution over $D_V$ on $\tilde{\mathbf{Q}}$ is finally formulated as:
\begin{equation}
\resizebox{0.88\linewidth}{!}{$
	\displaystyle 
	\begin{aligned}
	\ln p(\tilde{\mathbf{Q}}|D_V)&= \ln \prod_{(i,a,b)\in D_V} p(a >^+_i b|\tilde{\bm{Q}}_i)p(\tilde{\bm{Q}}_i)\\
	&=\sum_{i=1}^M\sum_{a=1}^K\sum_{b=1}^K I(a >_i^+ b)\ln\frac{1}{1+e^{-\lambda(\tilde{Q}_{ia}-\tilde{Q}_{ib})}}\\ 
	&+ C-\sum_{i=1}^M\sum_{j=1}^K \frac{\tilde{Q}_{ij}^2}{2\sigma^2},
	\end{aligned}
	$}
\label{eq:q_refining_loss}
\end{equation}
where $C$ is a constant that can be ignored during optimization. Before using $\tilde{\mathbf{Q}}$ in NeuralCDM, we need to restrict its elements to the range $(0,1)$, and set elements of concepts unlabeled or not predicted to 0. Thus, $Sigmoid(\tilde{\mathbf{Q}}) \circ \mathbf{M}$ is used to replace $\mathbf{Q}$ in NeuralCDM, where $\mathbf{M} \in \{0,1\}^{M\times K}$ is a mask matrix, and $M_{ij}=1$ if $j \in V^k_i$ or $Q_{ij}=1; M_{ij}=0$ otherwise. $\tilde{\mathbf{Q}}$ is trained together with the cognitive diagnostic model, thus the loss function is:
\begin{equation}
loss = -\ln p(\tilde{\mathbf{Q}}|D_V) + loss_{CDM}.
\end{equation}

\subsection{Generality of NeuralCD} \label{Generality}
In this subsection we show that NeuralCD is a general framework which can cover many traditional cognitive diagnostic models. Using Eq.~\eqref{eq:first_layer} as the first layer, we now show the close relationship between NeuralCD and traditional models, including MF, IRT and MIRT. 

\noindent\textbf{MF.} $\bm{Q}_e$ and $\bm{h}^s$ can be seen as exercise and student latent trait vectors respectively in MF. By setting $\bm{h}^{diff}\equiv\bm{0}$ and $h^{disc}\equiv 1$, the output of the first layer is $\bm{x}=\bm{Q}_e \circ \bm{h}^s$. Then in order to work like MF (i.e., $y=\bm{Q}_e \cdot \bm{h}^s$), all the rest of layers need to do is to sum up the values of each entry in $\bm{x}$, which is easy to achieve. Monotonicity assumption is not applied in MF approaches.

\noindent\textbf{IRT.} Take the typical formation of IRT $y={\rm Sigmoid}((h^s - h^{diff}) \times h^{disc})$ as example. Set $Q_e \equiv 1$, and let $\bm{h}^s$ and $\bm{h}^{diff}$ be unidimensional, the output of the first layer is $x=(h^s - h^{diff})\times h^{disc}$, followed by a Sigmoid activation function. Monotonicity assumption is achieved by limiting $h^{disc}$ to be positive. Other variations of IRT (e.g., $y'=C+(1-C)y$ where $C$ is guessing parameter) can be realized with a few changes.

\noindent\textbf{MIRT.} One direct extension from IRT to MIRT is to use multidimensional latent trait vectors of exercises and student. Here we take the typical formation proposed in~\cite{adams1997multidimensional} as example:
\begin{equation}
y=\frac{{\rm e}^{\mathbf{Q}_e \cdot \bm{h}^{s} - d_e}}{1 + {\rm e}^{\mathbf{Q}_e \cdot \bm{h}^{s} - d_e}}.
\end{equation}
Let $h^{disc}\equiv 1$, the output of the first layer given by Eq.~\eqref{eq:first_layer} is $\bm{x}=\mathbf{Q}_e \circ (\bm{h}^{s} - \bm{h}^{diff})$. By Setting
$
\mathbf{W}_1=\left[
\begin{matrix}
1&1&\cdots&1
\end{matrix}\right], \bm{b}_1=\bm{0}
$ 
and $\phi(x)=x$ in Eq.~\eqref{eq:full_layer1}, we have $f_1=\mathbf{Q}_e \cdot \bm{h}^{s} - d_e$ (where $d_e = \bm{Q}_e \cdot \bm{h}^{diff}$). All the rest of the layers need to do is to approximate the function $g(f_1)=1-{\rm Sigmoid}(f_1)$, which can be easily achieved with two more layers. Monotonicity assumption can be realized if each entry of $\bm{Q}_e$ is restricted to be positive.

\subsection{Discussion}
We have introduced the details of NeuralCD framework and showed special cases of it. It's necessary to point out that the student's proficiency vector $F^s$ and exercise's knowledge relevancy vector $F^{kn}$ are basic factors needed in NeuralCD framework. Additional factors such as exercise discrimination can be integrated into if reasonable. The formation of the first interactive layer is not limited, but it's better to contain the term $F^s \circ F^{kn}$ to ensure that each dimension of $F^s$ corresponds to a specific knowledge concept. The positive full connection is only one of the strategies that implement monotonicity assumption. More sophisticated network structures can be designed as the interaction layers. For example, recurrent neural network may be used to capture the time characteristics of the student's learning process.

\section{Experiments}
We first compare our NeuralCD models with some baselines on the student performance prediction task. Then we make some interpretation assessments of the models.

\subsection{Dataset Description} We use two real-world datasets in the experiments, i.e., Math and ASSIST. Math dataset supplied by iFLYTEK Co., Ltd. is collected from the widely-used online learning system Zhixue\footnote{https://www.zhixue.com}, which contains mathematical exercises and logs of high school examinations. ASSIST (ASSISTments 2009-2010 "skill builder") is an open dataset collected by the ASSISTments online tutoring systems~\cite{feng2009addressing}, which only provides student response logs and knowledge concepts\footnote{https://sites.google.com/site/assistmentsdata/home/assistment-2009-2010-data/skill-builder-data-2009-2010}. We choose the public corrected version that eliminates the duplicated data issue proposed by previous work~\cite{xiong2016going}. Table~\ref{tab:dataset_summary} summarizes basic statistics of the datasets.

We filter out students with less than 30 and 15 response logs for Math and ASSIST respectively to guarantee that each student has enough data for diagnosis. Therefore for dataset Math, we got 2,507 exercises with 497 knowledge concepts for diagnostic network, and the remaining exercises with knowledge concepts not appearing in logs are used for the Q-matrix refining part of NeuralCDM+. We perform a 80\%/20\% train/test split of each student's response log. As for ASSIST, we divide the response logs in the same way with Math, but NeuralCDM+ is not evaluated on this dataset as exercise text is not provided. All models are evaluated with 5-fold cross validation.

Students' knowledge proficiencies are stable in Math as the dataset is composed of logs from examinations. However, a student's proficiency on a knowledge concept may change as he will be continually given exercises of that concept until meeting certain criterion (e.g., answering 3 relevant exercises correctly in a row). To analyze whether static models (e.g., NeuralCD models and static traditional models) are suitable to apply on ASSIST, we compare two metrics between Math and ASSIST. The first metric is the average amount of logs that each student toke for each knowledge concept:
%\begin{equation}
%{\rm AVG}_{\#log} = \frac{\sum_i^N\sum_j^K Log(i,j)}{\sum_i^N\sum_j^K I(Log(i,j)>0)},
%\end{equation}
\begin{equation}
\resizebox{0.6\linewidth}{!}{$
	\displaystyle 
	{\rm AVG}_{\#log} = \frac{\sum_i^N\sum_j^K Log(i,j)}{\sum_i^N\sum_j^K I(Log(i,j)>0)},
	$}
\end{equation}
where $Log(i,j)$ is the amount of exercises student $s_i$ answered that related to knowledge concept $k_j$. Further, another metric is the mean standard deviation of scores $r_{ij}$ that $Log(i,j)>1$ as:
\begin{equation}
%\displaystyle
{\rm STD}_{\#log>1} = \mathop{\rm mean}\limits_{s_i \in S}(\mathop{\rm mean}\limits_{\substack{k_j \in K_n,\\ Log(i,j)>1}}(std_{ij})),
\end{equation}
where $std_{ij}$ is the standard deviation of scores that student $s_i$ got for exercises related to knowledge concept $k_j$. As the results showed in Table~\ref{tab:dataset_summary}, although ASSIST has a much larger ${\rm AVG}_{\#log}$ than Math, their ${\rm STD}_{\#log>1}$ are close. Therefore, it is reasonable to assume that the knowledge states of students in ASSIST are also stable, and our static NeuralCD models and baselines are applicable for both dataset. There will be more discussions in Model Interpretation.
\begin{table}
	\centering
	\caption{Dataset summary.}
	\label{tab:dataset_summary}
	%\begin{tabular}{p{4cm}p{1.5cm}<{\raggedleft}p{1.5cm}<{\raggedleft}}
	\begin{tabular}{lrr}
		\toprule
		Dataset & Math & ASSIST \\
		\midrule
		\#Students  & 10,268 & 4,163 \\
		\#Exercises & 917,495 & 17,746 \\
		\#Knowledge concepts & 1,488 & 123 \\
		\#Response logs & 864,722 & 324,572 \\
		\#Knowledge concepts per exercise & 1.53 & 1.19 \\
		${\rm AVG}_{\#log}$ & 2.28 & 8.05 \\
		${\rm STD}_{\#log>1}$ & 0.305 & 0.316 \\
		\bottomrule
	\end{tabular}
\end{table}

\subsection{Experimental Setup} The dimensions of the full connection layers (Eq.~\eqref{eq:full_layer1} $\sim$~\eqref{eq:output_layer}) are 512, 256, 1 respectively, and Sigmoid is used as activation function for all of the layers. We set hyperparameters $\lambda=0.1$ (Eq.~\eqref{eq:q_logistic}) and $\sigma=1$ ( Eq.~\eqref{eq:q_refining_loss}). For $k$ in top-k knowledge concepts selecting, we use the value that make the predicting network reach 0.85 recall. That is, in our experiment, $k=20$. We initialize the parameters with {\it Xavier} initialization~\cite{glorot2010understanding}, which fill the weights with random values sampled from $\mathcal{N}(0, std^2)$, where $std=\sqrt{\frac{2}{n_{in}+n_{out}}}$. $n_{in}$ is the number of neurons feeding into the weights, and $n_{out}$ is the number of neurons the results is fed to. %, and use AdamOptimizer with learning rate 0.0005.

The CNN architecture we use in NeuralCDM+ contains 3 convolutional layers followed by a full connection output layer. MaxPooling are used after 1st and 3rd convolutional layers. The channels of convolutional layers are 400, 200, 100, and kernel sizes are set to 3, 4, 5 respectively. We adopt ReLu activation function for convolution layers and Sigmoid for the output layer. Multi-label binary cross entropy is used as loss function for training the CNN.

To evaluate the performance of our NeuralCD models, we compare them with previous approaches, i.e., DINA, IRT, MIRT and PMF. All models are implemented by PyTorch using Python, and all experiments are run on a Linux server with four 2.0GHz Intel Xeon E5-2620 CPUs and a Tesla K20m GPU. %For fairness, all models are tuned to have the best performance.

\begin{table*}[htb]
	\centering
	\caption{Experimental results on student performance prediction.}
	\label{tab:exp_result}
	\begin{tabular}{m{1.8cm}m{1.5cm}m{1.5cm}m{1.5cm}m{1.5cm}m{1.5cm}m{1.5cm}}
		\toprule
		& \multicolumn{3}{c}{Math} & \multicolumn{3}{c}{ASSIST} \\
		\cmidrule(r){2-4}  \cmidrule(r){5-7}
		Model & Accuracy & RMSE & AUC & Accuracy & RMSE & AUC \\
		\midrule
		DINA & 0.593$\pm$.001 & 0.487$\pm$.001 & 0.686$\pm$.001 & 0.650$\pm$.001 & 0.467$\pm$.001 & 0.676$\pm$.002 \\
		IRT & 0.782$\pm$.002 & 0.387$\pm$.001 & 0.795$\pm$.001 & 0.674$\pm$.002 & 0.464$\pm$.002 & 0.685$\pm$.001 \\
		MIRT & 0.793$\pm$.001 & 0.378$\pm$.002 & 0.813$\pm$.002 & 0.701$\pm$.002 & 0.461$\pm$.001 & 0.719$\pm$.001 \\
		PMF & 0.763$\pm$.001 & 0.407$\pm$.001 & 0.792$\pm$.002 & 0.661$\pm$.002 & 0.476$\pm$.001 & 0.732$\pm$.001 \\
		NeuralCDM & 0.792$\pm$.002 & 0.378$\pm$.001 & 0.820$\pm$.001 & \textbf{0.719$\pm$.008} & \textbf{0.439$\pm$.002} & \textbf{0.749$\pm$.001} \\
		NeuralCDM+ & \textbf{0.804$\pm$.001} & \textbf{0.371$\pm$.002} & \textbf{0.835$\pm$.002} & - & - & - \\
		\bottomrule
	\end{tabular}
\end{table*}
\begin{figure*}[ht]
	\begin{minipage}[h]{0.47\textwidth}
		\centering
		\includegraphics[scale=0.49]{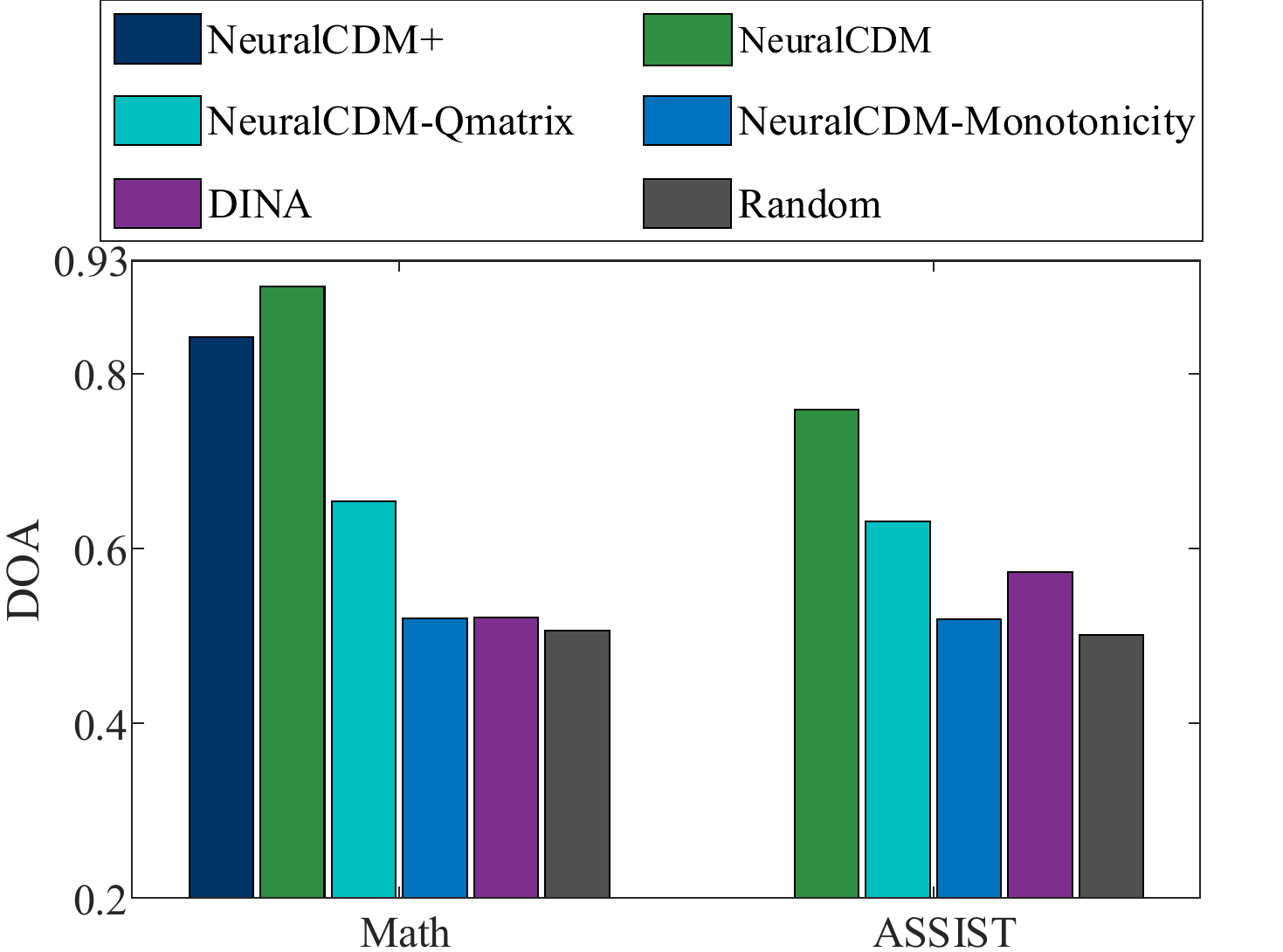}
		\caption{DOA results of models. In NeuralCD models, there is clear correspondence between entries in $\bm h^s$ and knowledge concepts, thus their diagnosis results have high DOA. Removing Q-matrix or monotonicity assumption would reduce the performance.}
		\label{fig:doa_result}
	\end{minipage}\hspace{9mm}
	\begin{minipage}[h]{0.47\textwidth}
		\centering
		\includegraphics[scale=0.38]{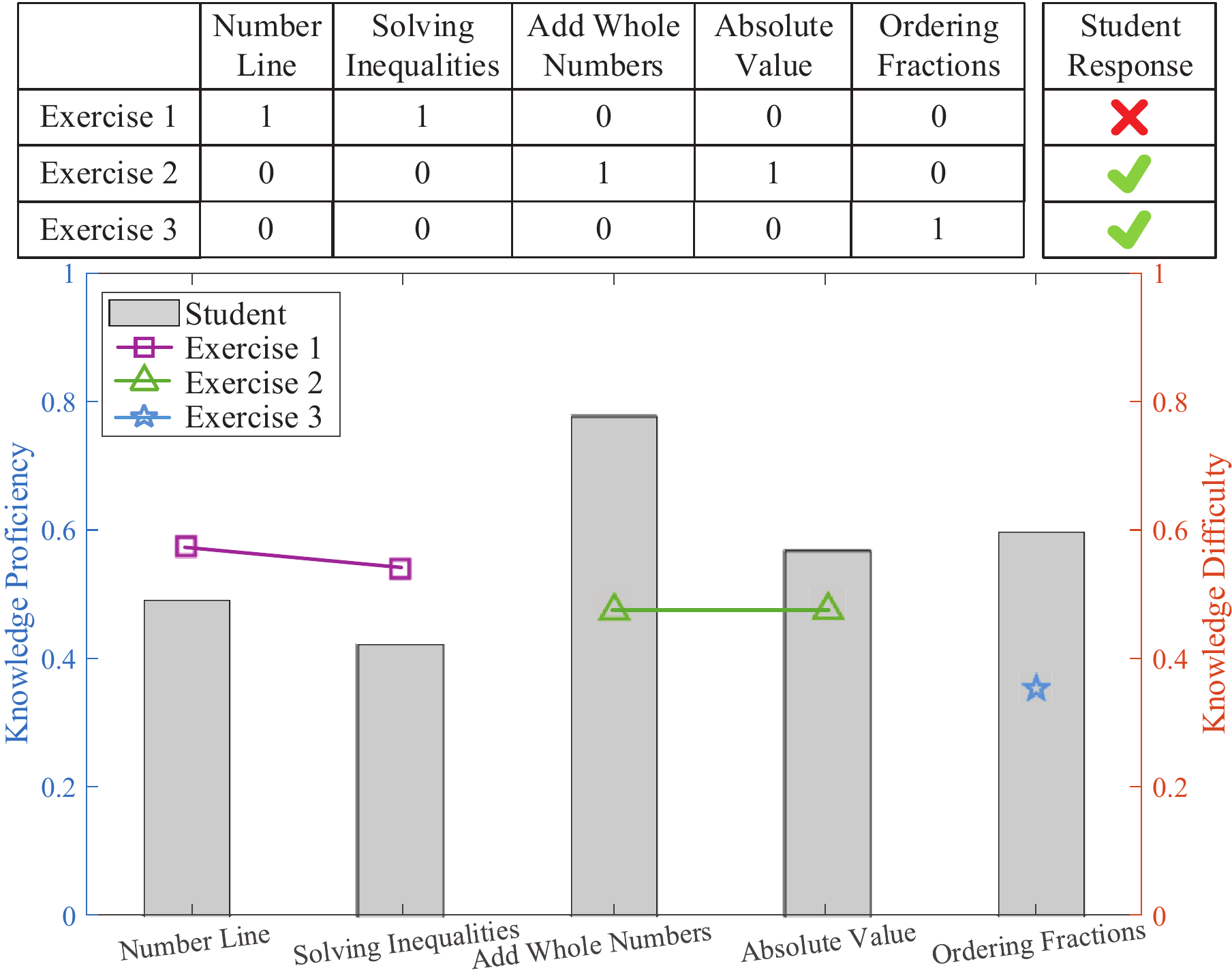}
		\caption{Diagnosis example of a student in ASSIST. The upper part is the Q-matrix of 3 exercises and corresponding response logs. The lower part shows the diagnosed student's knowledge proficiencies (bars) and  knowledge difficulties of each exercise (points).}
		\label{fig:case_study}
	\end{minipage}
\end{figure*}

\subsection{Experimental Results}
\subsubsection{Student Performance Prediction}
The performance of a cognitive diagnosis model is difficult to evaluate as we can't obtain the true knowledge proficiency of students. As diagnostic result is usually acquired through predicting students' performance in most works, performance on these prediction tasks can indirectly evaluate the model from one aspect~\cite{liu2018fuzzy}. Considering that all the exercises we used in our data are objective exercises, we use evaluation metrics from both classification aspect and regression aspect, including accuracy, RMSE (root mean square error)~\cite{pei2018group} and AUC (area under the curve)~\cite{bradley1997use}. 

Table~\ref{tab:exp_result} shows the experimental results of all models on student performance prediction task. The error bars after '$\pm$' is the standard deviations of 5 evaluation runs for each model. From the table, we can observe that NeuralCD models outperform almost all the other baselines on both datasets, indicating the effectiveness of our framework. In addition, the better performance of NeuralCDM+ over NeuralCDM proves that the Q-matrix refining method is effective, and also demonstrates the importance of fine estimated knowledge relevancy vectors for cognitive diagnosis.

\subsubsection{Model Interpretation}
To assess the interpretability of NeuralCD framework (i.e., whether the diagnostic result is reasonable), we further conduct several experiments. 

Intuitively, if student $a$ has a better mastery on knowledge concept $k$ than student $b$, then $a$ is more likely to answer exercises related to $k$ correctly than $b$ ~\cite{chen2017tracking}. We adopt Degree of Agreement (DOA)~\cite{pirotte2007random} as the evaluation metric of this kind of ranking performance. For knowledge concept $k$, $DOA(k)$ is formulated as:
\begin{equation}
\resizebox{0.88\linewidth}{!}{$
	\displaystyle 
	DOA(k)=\frac{1}{Z} \sum_{a=1}^{N}\sum_{b=1}^{N} \delta(F_{ak}^s,F_{bk}^s) \sum_{j=1}^{M} I_{jk} \frac{J(j,a,b) \land \delta(r_{aj},r_{bj})}{J(j,a,b)},
	$}
\end{equation}
where $Z = \sum_{a=1}^{N}\sum_{b=1}^{N} \delta(F_{ak}^s,F_{bk}^s)$.
$F_{ak}^s$ is the proficiency of student $a$ on knowledge concept $k$. $\delta(x,y)=1$ if $x>y$ and $\delta(x,y)=0$ otherwise. $I_{jk}=1$ if exercise $j$ contains knowledge concept $k$ and $I_{jk}=0$ otherwise. $J(j,a,b)=1$ if both student $a$ and $b$ did exercise $j$ and $J(j,a,b)=0$ otherwise. We average $DOA(k)$ on all knowledge concepts to evaluate the quality of diagnostic result (i.e., knowledge proficiency acquired by models).

Among traditional models, we only compare with DINA, since for IRT, MIRT and PMF, there are no clear correspondence between their latent features and knowledge concepts. Besides, we conduct experiments on two reduced NeuralCDM models. In the first reduced model (denoted as NeuralCDM-Qmatrix), knowledge relevancy vectors are estimated during unsupervised training instead of getting from Q-matrix. While in another reduced model (denoted as NeuralCDM-Monotonocity), monotonicity assumption is removed by eliminating the positive restriction on the full connection layers. These two reduced models are used to demonstrate the importance of fine-estimated knowledge relevancy vector and monotonicity assumption respectively. Furthermore, we conduct an extra experiment in which students' knowledge proficiencies are randomly estimated, and compute the DOA for comparison.

Figure~\ref{fig:doa_result} presents the experimental results. From the figure we can observe that DOAs of NeuralCDM and NeuralCDM+ are significantly higher than baselines, which proves that knowledge proficiencies diagnosed by them are reasonable. The DOAs of NeuralCDM-Qmatrix and NeuralCDM-Monotonicity are much lower than NeuralCDM, which indicates that both information from Q-matrix and monotonicity assumption are important for getting interpretable diagnosis results (knowledge proficiency vectors). DOA of DINA is slightly higher than Random due to the use of Q-matrix. Besides, NeuralCDM performs much better on Math than on ASSIST. This is mainly due to the contradictions in logs, i.e., a student may answer some exercises containing knowledge concept $k_j$ correctly while others containing $k_j$ wrong (reasons may be the change of knowledge proficiency, or other knowledge concepts contained by the exercises). As showed in Table~\ref{tab:dataset_summary}, ASSIST has much larger ${\rm AVG}_{\#log}$ and slightly higher ${\rm STD}_{\#log>1}$ than Math dataset, which makes more contradictions in logs. Longer logs with more contradictions would decrease DOA.

\subsubsection{Case Study.}
Here we present an example of a student's diagnostic result of NeuralCDM on dataset ASSIST in Figure~\ref{fig:case_study}. The upper part of Figure~\ref{fig:case_study} shows the Q-matrix of three exercises on five knowledge concepts and the response of a student to the exercises. The bars in the underneath subfigure represent the student's proficiency on each knowledge concept. The lines with different colors and markers represent the knowledge difficulties of the three exercises (for clarity, we only present difficulties of relevant knowledge concepts for each exercise). We can observe from the figure that the student is more likely to response correctly when his proficiency satisfies the requirement of the exercise. For example, exercise 3 requires the mastery of '{\it Ordering Fraction}' and corresponding difficulty is 0.35. The student's proficiency on '{\it Ordering Fraction}' is 0.60, which is higher than required, thus he answered it correctly. Both knowledge difficulty ($\bm{h}^{diff}$) and knowledge proficiency ($\bm{h}^s$) in NeuralCDM are explainable as expected.

\subsection{Discussion.} From the above experiments, we can observe that NeuralCD models provide both accurate and interpretable results for cognitive diagnosis. 

There still some directions for future studies. First, we may make our effort to design a more efficient model for knowledge concept prediction, which would promote the performance of NeuralCDM+. Second, the positive restriction on neural network weights may limit the approximate ability, thus we would like to explore more flexible methods to satisfy the monotonicity assumption.
Third, since students' knowledge statuses change in many online self-learning circumstances, we would like to extend NeuralCD for dynamic cognitive diagnosis.

\section{Conclusion}
In this paper, we proposed a neural cognitive diagnostic framework, NeuralCD framework, for students' cognitive diagnosis. Specifically, we first discussed fundamental student and exercise factors in the framework, and placed a monotonicity assumption on the framework to ensure its interpretability. Then, we implemented a specific model NeuralCDM under the framework to show its feasibility, and further extended NeuralCDM by incorporating exercise text to refine Q-matrix. Extended experimental results on real-world datasets showed the effectiveness of our models with both accuracy and interpretability. We also showed that NeuralCD could be seen as the generalization of some traditional cognitive diagnostic models (e.g., MIRT). The structure of the diagnostic network in our work is designed intuitively. However, with the high flexibility and potential of neural network, we hope this work could lead to further studies.

\section{ Acknowledgments}
This research was partially supported by grants from the National Natural Science Foundation of China (Grants No. 61922073, 61672483, U1605251, 61727809), the Science Foundation of Ministry of Education of China \& China Mobile (No. MCM20170507), and the Iflytek joint research program.

%\small
%\fontsize{9.8pt}{10.8pt} \selectfont
\fontsize{9.3pt}{10.3pt} \selectfont
\bibliographystyle{aaai}
\bibliography{1258-NeuralCD}

\end{document}